\documentclass[10pt,twocolumn,letterpaper]{article}

\usepackage[pagenumbers]{cvpr}           %

\usepackage{graphicx}
\usepackage{amsmath}
\usepackage{amssymb}
\usepackage{booktabs}
\usepackage[accsupp]{axessibility}  %

\usepackage[pagebackref,breaklinks,colorlinks]{hyperref}

\usepackage[capitalize]{cleveref}
\crefname{section}{Sec.}{Secs.}
\Crefname{section}{Section}{Sections}
\Crefname{table}{Table}{Tables}
\crefname{table}{Tab.}{Tabs.}

\def\cvprPaperID{9497} %
\def\confName{CVPR}
\def\confYear{2023}

\usepackage{dsfont}
\usepackage{amsthm}
\usepackage{overpic}
\usepackage{contour}
\contourlength{1pt}
\usepackage{xspace}
\usepackage{sidecap}
\usepackage{rotating}
\usepackage[normalem]{ulem}
\usepackage{booktabs} %
\usepackage{caption}
\usepackage{subcaption}
\usepackage{microtype}
\usepackage{booktabs}
\usepackage{enumitem}
\usepackage{amsthm}
\usepackage{wrapfig}
\usepackage{multirow}
\usepackage{tabularx}
\usepackage{nicefrac}
\usepackage{algorithm}
\usepackage{algorithmicx}
\usepackage{algpseudocode}
\usepackage{chngcntr}
\usepackage{apptools}
\usepackage[super]{nth}
\AtAppendix{\counterwithin{lemma}{section}}

\usepackage{ tipa }
\usepackage{adjustbox }

\newcommand{\RR}{\mathbb{R}}

\newcommand{\PP}{\mathcal{P}}

\newlength{\indentlaenge}
\newlength{\mylength}
\newlength{\mylengthzwei}
\def\cput(#1,#2)#3{\put(#1,#2){\hbox to 0pt{\hss{#3}\hss}}}
\def\lput(#1,#2)#3{\put(#1,#2){\hbox to 0pt{\hss{#3}}}}
\def\rput(#1,#2)#3{\put(#1,#2){\hbox to 0pt{{#3}\hss}}}

\begin{document}

\title{OReX: Object Reconstruction from Planar Cross-sections Using Neural Fields}

\author{Haim Sawdayee\\
The Blavatnik School of Computer Science\\
Tel Aviv University\\
{\tt\small haimsawdayee@mail.tau.ac.il}
\and
Amir Vaxman\\
School of Informatics\\
The University of Edinburgh\\
{\tt\small avaxman@inf.ed.ac.uk}
\and
Amit H. Bermano\\
The Blavatnik School of Computer Science\\
Tel Aviv University\\
{\tt\small amberman@tauex.tau.ac.il}
}

\twocolumn[\maketitle\vspace{-3em}
\begin{center}
    \includegraphics[width=0.95\textwidth]{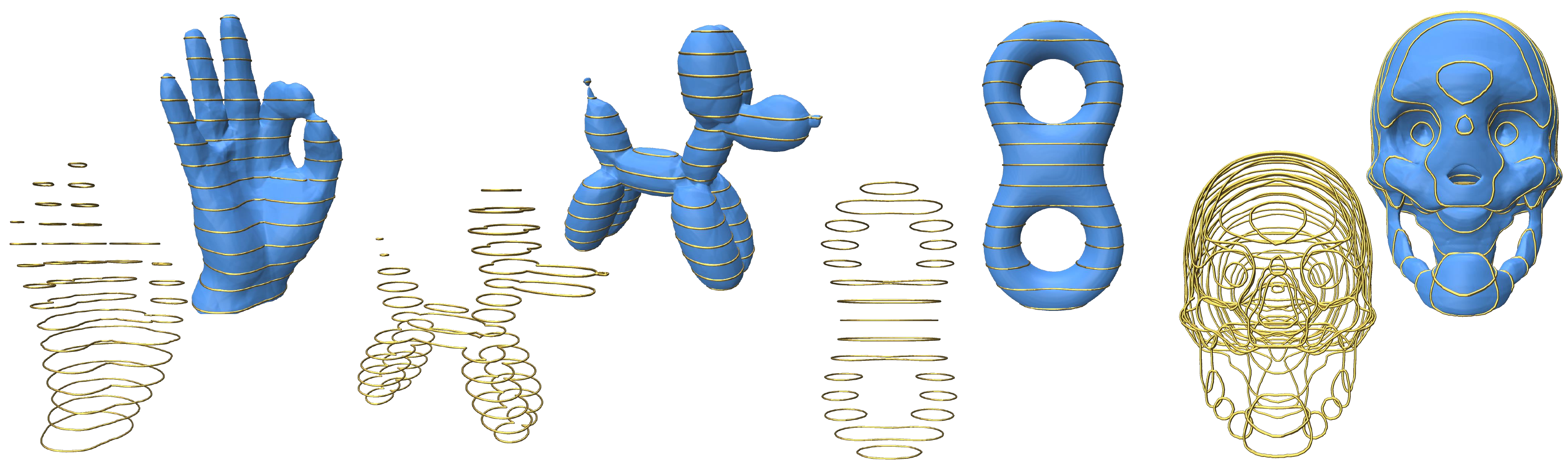}
\end{center}\vspace{-1.5em}

\captionof{figure}{OReX reconstructs smooth 3D shapes (right) from input planar cross-sections (left). Our neural field-based prior allows smooth interpolation between slices while respecting high-frequency features and self-similarities.}
\label{fig:teaser}

\bigbreak]

\begin{abstract}
   Reconstructing 3D shapes from planar cross-sections is a challenge inspired by downstream applications like medical imaging and geographic informatics. The input is an in/out indicator function fully defined on a sparse collection of planes in space, and the output is an interpolation of the indicator function to the entire volume. Previous works addressing this sparse and ill-posed problem either produce low quality results, or rely on additional priors such as target topology, appearance information, or input normal directions. In this paper, we present OReX, a method for 3D shape reconstruction from slices alone, featuring a Neural Field as the interpolation prior. A modest neural network is trained on the input planes to return an inside/outside estimate for a given 3D coordinate, yielding a  powerful prior that induces smoothness and self-similarities. The main challenge for this approach is high-frequency details, as the neural prior is overly smoothing. To alleviate this, we offer an iterative estimation architecture and a hierarchical input sampling scheme that encourage coarse-to-fine training, allowing the training process to focus on high frequencies at later stages. In addition, we identify and analyze a ripple-like effect stemming from the mesh extraction step. We mitigate it by regularizing the spatial gradients of the indicator function around input in/out boundaries during network training, tackling the problem at the root.  
   Through extensive qualitative and quantitative experimentation, we demonstrate our method is robust, accurate, and scales well with the size of the input. We report state-of-the-art results compared to previous approaches and recent potential solutions, and demonstrate the benefit of our individual contributions through analysis and ablation studies.
\end{abstract}
\footnote{Code and data available at \url{https://github.com/haimsaw/OReX}}

\section{Introduction}
\label{sec:intro}

Reconstructing a 3D object from its cross-sections is a long-standing task. It persists in fields including medical imaging, topography mapping, and manufacturing. The typical setting is where a sparse set of arbitrary planes is given, upon which the ``inside'' and ``outside'' regions of the depicted domain are labeled, and the entire shape in 3D is to be estimated (see Fig.~\ref{fig:teaser}). This is a challenging and ill-posed problem, especially due to the sparse and irregular nature of the data. Classical approaches first localize the problem by constructing an arrangement of the input planes, and then introduce a local regularizer that governs the interpolation of the input to within each cell. While sound, these approaches typically involve simplistic regularization functions, that only interpolate the volume within a cell bounded by the relevant cross-sections; as a consequence, they introduce over-smoothed solutions that do not respect features. In addition, finding a cellular arrangement of planes is a computationally-heavy procedure, adding considerable complexity to the problem and rendering it quickly infeasible for large inputs (see Sec.~\ref{sec:results}). As we demonstrate (Sec.~\ref{sec:results}), recent approaches that reconstruct a mesh from an input point cloud are not well suited to our setting, as they assume a rather dense sampling of the entire shape. In addition, these methods do not consider the information of an entire cross-sectional plane, but rather only on the shape boundary.

In this paper, we introduce \emph{OReX}---a reconstruction approach based on neural networks that estimates an entire shape from its cross-sectional slices. Similar to recent approaches, the neural network constitutes the prior that extrapolates the input to the entire volume. Neural networks in general have already been shown to inherently induce smoothness~\cite{ma2021neural}, and self-similarities~\cite{hanocka2020point2mesh}, allowing natural recurrence of patterns. Specifically, we argue that Neural Fields~\cite{xie2022neural} are a promising choice for the task at hand. Neural Fields represent a 3D scene by estimating its density and other local geometric properties for every given 3D coordinate. They are typically trained on 2D planar images, and are required to complete the entire 3D scene according to multi-view renderings or photographs. This neural representation is differentiable by construction, and hence allows native geometric optimization of the scene, realized via training. We pose the reconstruction problem as a classification of space into ``in'' and ``out'' regions, which are known for the entire slice planes, and thus generate the indicator function which its decision boundary is the output shape.

The main challenge with applying neural fields to this problem is high-frequency details. Directly applying established training schemes \cite{mildenhall2021nerf} shows strong spectral bias, yielding overly smoothed results and other artifacts (Fig.~\ref{fig:basic-nerf}). Spectral bias is a well-known effect, indicating that higher frequency is effectively learned slower \cite{rahaman2019spectral}. To facilitate effective high-frequency learning, avoiding the shadow cast by the low frequency, we introduce two alterations.  First, we sample the planar data mostly around the inside/outside transition regions, where the frequencies are higher. This sampling is further ordered from low to high-frequency regions (according to the distance from the inside/outside boundary), to encourage a low-to-high-frequency training progression. In addition, we follow recent literature and allow the network to iteratively infer the result, where later iterations are responsible for finer, higher-frequency corrections \cite{alaluf2021restyle,song2020denoising}.

Finally, we consider another high-frequency artifact, also found in other neural-field-based works \cite{erler2020points2surf}. The desired density (or indicator) function dictates a sharp drop in value at the shape boundary. This is contradictory to the induced neural prior, causing sampling-related artifacts in the downstream task of mesh extraction (Sec.~\ref{subsec:loss-function-and-inference}). To alleviate this, we penalize strong spatial gradients around the boundary contours. This enforces a smoother transition between the in and out regions, allowing higher-quality mesh extraction.  

As we demonstrate (see Fig.~\ref{fig:teaser}), our method yields state-of-the-art reconstructions from planar data, both for woman-made and organic shapes.  The careful loss and training schemes are validated and analyzed through quantitative and qualitative experimentation. Our method is arrangement-free, and thus both interpolates all data globally, avoiding local artifacts, and scales well to a large number of slices (see Sec.~\ref{sec:results}).

\section{Related work}
\label{sec:related}

\paragraph{Reconstruction from cross-sections}

Reconstruction from planar cross-section has been a long-standing challenge in geometry processing and computational geometry. The problem was mostly studied in the parallel planes setting (e.g.,\cite{barequet1994piecewise, boissonnat2007shape, bajaj1996arbitrary}), where the 3D object was reconstructed per two such ``keyframes''. Later work offered general solutions for any orientation or distribution of planes \cite{bermano2011online, liu2008surface}. The general approach was based on tessellating the planes into convex cells by the planes of the arrangement and reconstructing the object in each cell, by defining some interpolant inside it. These methods suffer from several issues: the construction of the arrangement is computationally expensive (with at least cubic asymptotic times), and the local reconstruction introduces continuity artifacts (see Sec.~\ref{sec:results}). Furthermore, the interpolants were usually designed for smoothness or proximity, and failed to capture more global trends in the geometry of the slices, such as twists or extrusions (see Fig.~\ref{fig:comparisons_visual}). Some works provide topological guarantees \cite{zou2015topology,huang2017topology} or limited the solution to given templates \cite{holloway2006template}, but do not provide a general solution without these priors. Our work is arrangement-free and provides a flexible high-parametric model, using neural networks, to capture the details of the reconstructed object.

\begin{figure*}
\centering
\includegraphics[width=0.99\textwidth]{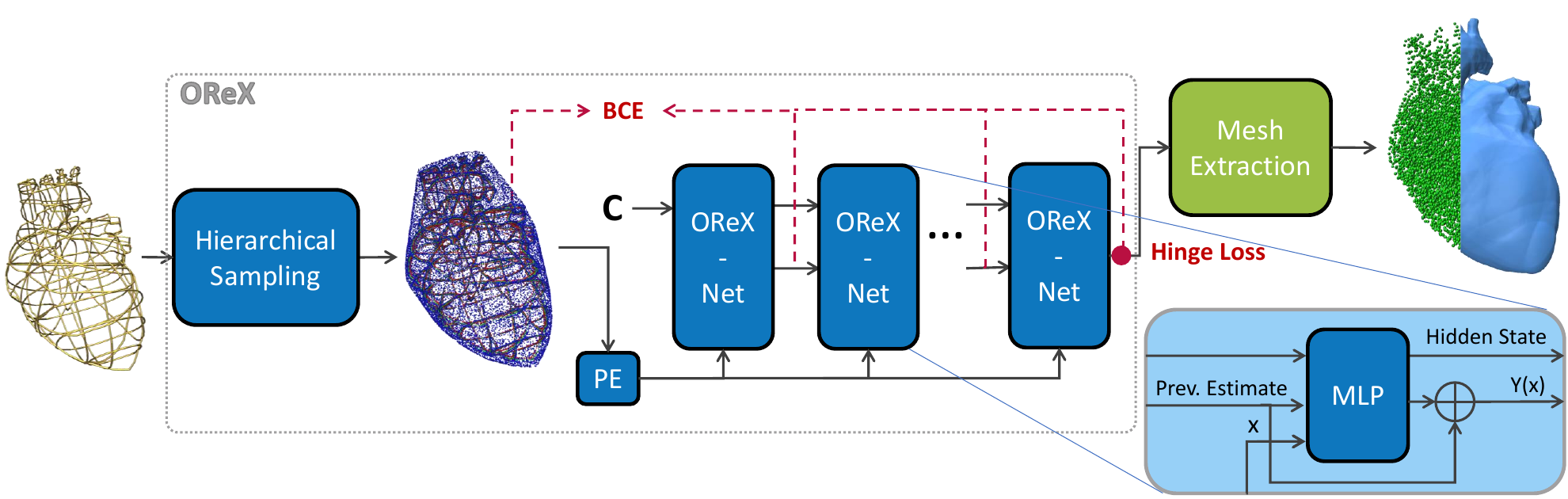}
\caption{OReX method overview. The input (left) is a set of planar cross-sections with inside/outside information on them. The planes are sampled into points in a hierarchical scheme, and are fed to our iterative Neural Field for training. After training, the final shape is estimated using an off-the-shelf mesh extractor.}
\label{fig:pipeline}
\end{figure*}

\begin{figure}
    \centering
     \includegraphics[width=0.45\textwidth]{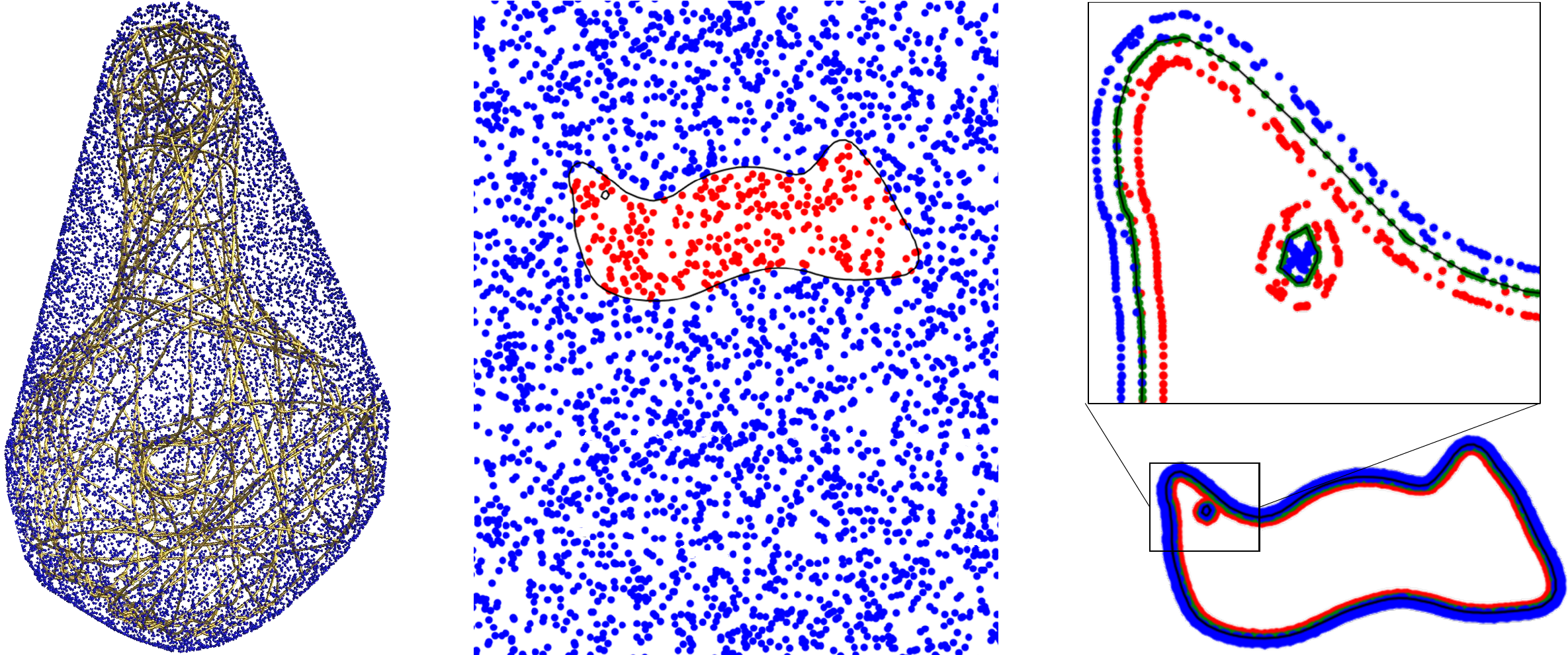}
    \caption{We train OReX on three types of sampling distributions of the input: A scaled-up version of the 3D convex hull bounds the reconstruction volume (left); Uniform sampling within each plane helps stabilize the learning (middle); most of the samples are concentrated where accuracy matters most---on and around the boundary contours on the input planes (right).}
    \label{fig:sampling_method}
\end{figure}

\paragraph{Neural reconstruction} Most recent reconstruction works employing neural networks address reconstruction from dense point clouds. Our problem could be cast as a point cloud reconstruction one, after sampling the planes. In this setting, the high-frequency problem is pronounced as in our case, and even more so since more information regarding the target shape is available. Almost all recent approaches employ an implicit representation of the shape. Some works populate 3D (sometimes adaptive) grids to handle fine-details \cite{chen2021multiresolution,chibane2020implicit,martel2021acorn,takikawa2021neural}, which is challenging to scale. Other works \cite{peng2020convolutional,boulch2022poco} perform local operations, which rely on a rather dense neighborhood. To alleviate this, a mesh can be directly optimized to match the input \cite{hanocka2020point2mesh}. Other neural fields have been employed as well in this context \cite{ma2021neural}, without additional information such as normals; see a recent survey for more approaches ~\cite{xie2022neural}. As we demonstrate (Sec.~\ref{sec:results}), this general line of works employs weaker priors, which do not fit the sparse nature of our problem well. 

Perhaps a more similar setting is the one of reconstruction from projections or photographs. Here the reconstructions are from 2D data as well, which is also of sparser nature, albeit using appearance information that we do not have. Literature in this field is deep and wide, with several surveys \cite{ham2019computer,gao2022nerf}, including the usage of neural fields. We argue our work is orthogonal to that listed here, as it can be plugged in to replace our basic neural-field baseline. We leave for future work to inspect the most performing architecture. Addressing the most similar problem setting, concurrent work by Ostonov et al.~\cite{ostonov2022cut} employs reinforcement learning and orthographic projections to ensure proper reconstruction. Except for requiring additional information during inference, they also require a training phase and are restricted to the domain trained on. Our approach, in contrast, requires no additional information besides the planes.

\section{Method}
\label{sec:method}

We next lay out the details for effective high-quality shape reconstruction from slices. Our approach is based on a neural field, \textit{OReXNet}. OReXNet outputs $Y(x)$, an extrapolation of the inside/outside indicator function for a query point $x \in \mathbb{R}^3$ (Sec.~\ref{subsec:problem-setting}). Given a single set of input cross-sections, the network undergoes training to approximate the target function on the input. After training is complete, $Y$ is sampled on the entire volume, and a resulting shape is extracted using the Dual Contouring (DC) approach~\cite{ju2002dual}. This pipeline is depicted in Fig.~\ref{fig:pipeline}.

As motioned in Sec.~\ref{sec:intro}, the main challenge in reconstruction quality is high-frequency details. The straightforward approach to our problem would be to train a neural field for the desired indicator function by uniformly sampling all planes, and subsequently training a network using a \emph{Binary Cross Entropy} (BCE) loss on all sampled points. Typical approaches also represent the input coordinates using Positional Encoding (PE) \cite{mildenhall2021nerf}. As it turns out, this approach yields overly smooth results failing even to interpolate the input (Fig.~\ref{fig:basic-nerf}). To improve reconstruction fidelity, and allow higher frequency details in the resulting shape, we introduce two alterations to the aforementioned training scheme. We present our hierarchical input sampling scheme in Sec.~\ref{subsec:sampling}, and our iterative-refinement architecture in Sec.~\ref{subsec:architecture}.

Finally, we also address a common artifact in implicit mesh extraction \cite{erler2020points2surf}. Repeating patterns can be seen that correlate to the sampling pattern of the mesh extraction phase, as demonstrated in Fig.~\ref{fig:hinge-loss}. We describe how we design the loss function to mitigate this artifact in Sec.~\ref{subsec:loss-function-and-inference}. See the Supplementary material and code for implementation details and hyper-parameter values. 

\begin{figure*}
    \centering
    \includegraphics[width=0.95\textwidth]{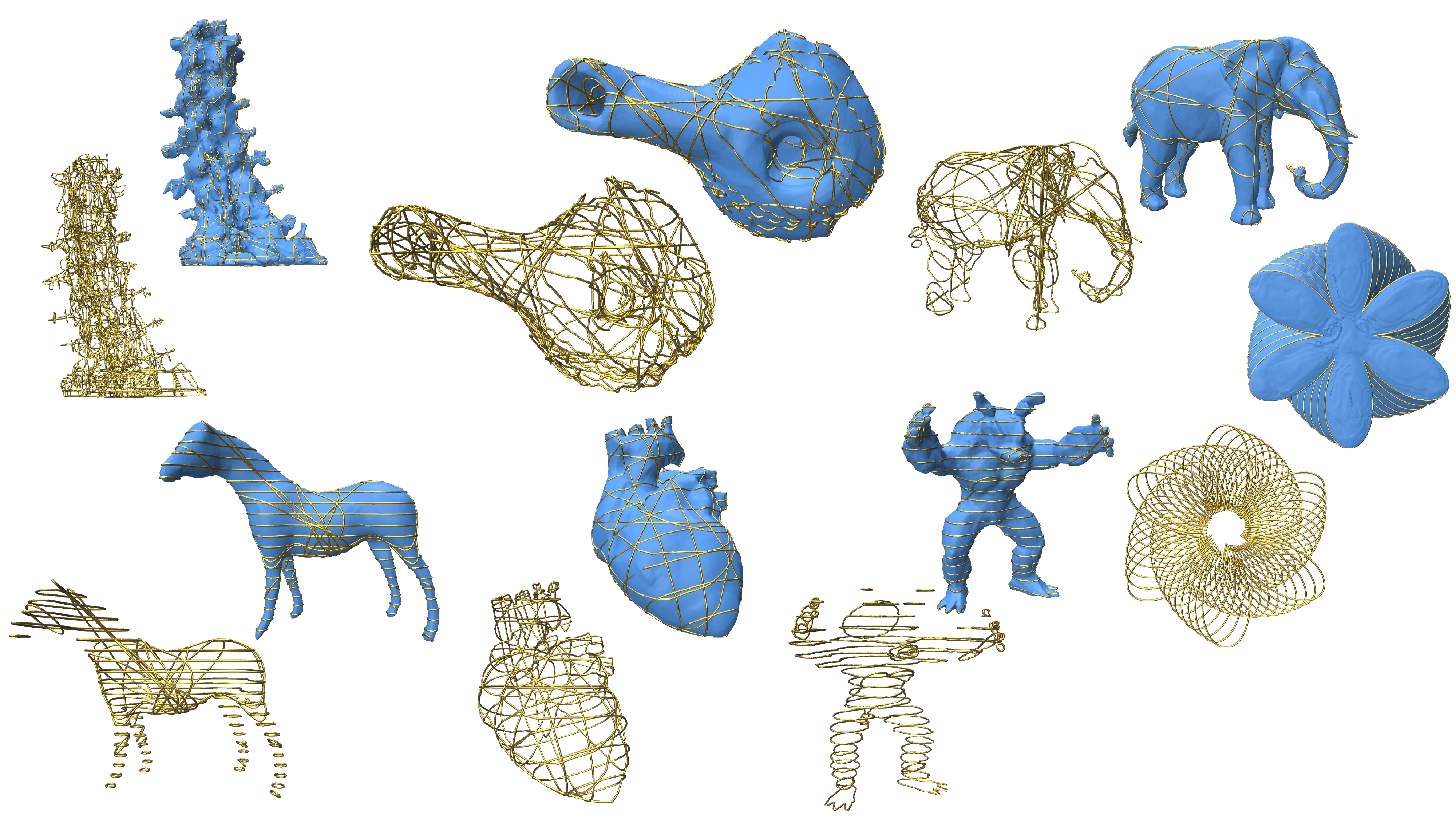}
    \caption{Results gallery. A qualitative demonstration of a collection of man-made and medical inputs. Zoomed-in viewing is recommended.}
    \label{fig:gallery}
\end{figure*}

\begin{figure}
    \centering
    \includegraphics[width=0.5\textwidth]{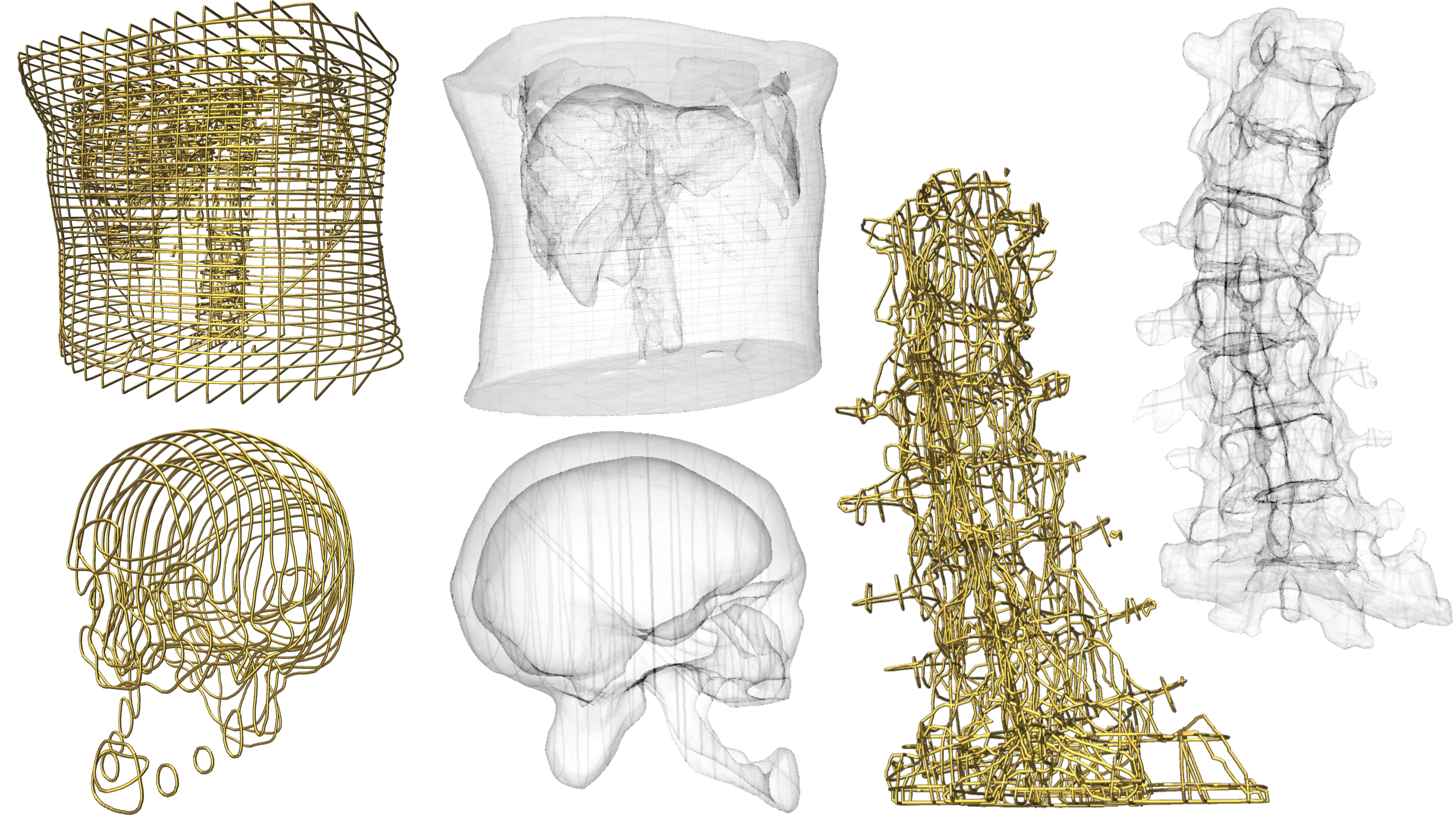}
    \caption{Additional qualitative results, where transparency  reveals the reconstruction of internal cavities and tunnels. Zoomed-in viewing is recommended.}
    \label{fig:xray}
\end{figure}

\subsection{Problem Setting}
\label{subsec:problem-setting}

We consider a set $\PP=\left\{P_1,\cdots, P_k\right\}$ of 2D planes embedded in $\RR^3$, with arbitrary offsets and orientations. 
Each plane $P_i$ contains an arbitrary set of (softly) non-intersecting oriented contours $C_{i} = \left\{c_{i,1},\cdots,c_{i,{l_i}}\right\}$ that consistently partition the plane into regions of ``inside'' and ``outside'' of an unknown domain $\Omega \subset \RR^3$ with boundary $\partial \Omega$ (Fig.~\ref{fig:pipeline}). The target output of our method is an indicator function $Y:\RR^3 \rightarrow \RR$, defining $\Omega$ as:
\begin{equation}
    Y(x) = \begin{cases}1\ & x \notin \Omega \\
    0 & x \in \Omega \\
    0.5 & x \in \partial \Omega
    \end{cases}
\end{equation}
In practice, we approximate $Y$ using a function $f$, such that $Y(x) \approx
 \sigma\left(f\left(x\right)\right) $, where $\sigma(z) = \frac{1} {1 + e^{-z}}$  is the sigmoid function.

\subsection{Input Sampling}
\label{subsec:sampling}

  Regularly or randomly sampling the input planes for training yields inaccurate and overly smooth results (see Fig.~\ref{fig:basic-nerf}). Instead, we define three types of relevant point distributions to sample from, as depicted in Fig.~\ref{fig:sampling_method}:
\begin{enumerate} 
    \item We bound the reconstructed volume by the 3D convex hull of all the input contours, and scale it up by $5\%$. We consequently sample it uniformly, where the sampled points are all outside.
    \item We compute the bounding box (aligned to the principal axes) of the contours in each plane, and sample it with uniform distribution.
    \item Most samples are taken around the boundaries of the contours, as these are the regions where accuracy matters most. We then sample every contour edge evenly, and further add off-surface points of varying distances for each edge sample. Off-surface locations are found by moving away from the contour on the plane (i.e., in the direction of the plane normal). We further sample off-surface points in a circle around each vertex. 
\end{enumerate}

Each sampling point $x_i$ is matched with a label $Y_i$ according to its position on the slice, and the pairs $\{x_i,Y_i\}$ constitute the input to our training. See supplementary material for the relevant hyperparameters.

\textbf{Frequency-Oriented Sampling.} In order to encourage better high-frequency learning, we sample the $3^{rd}$ type of points around a set of varying distances, from $0.1\%$ of the bounding box away from the contours to three orders of magnitude closer. In every epoch, we only use points sampled around three consecutive distance ranges. For early training iterations, we use the three largest distances, since further away points translate to lower frequency information about the shape. As the training progresses, and the lower frequencies are assumed to stabilize, we train with points that grow closer and closer to the actual contour, thus focusing the learning process on higher and higher frequencies (see Fig.~\ref{fig:no-refinment}). See supplementary material for more details and exact scheduling. %

\subsection{Architecture}
\label{subsec:architecture}

\begin{figure}[ht]
    \centering
    \includegraphics[width=0.47\textwidth]{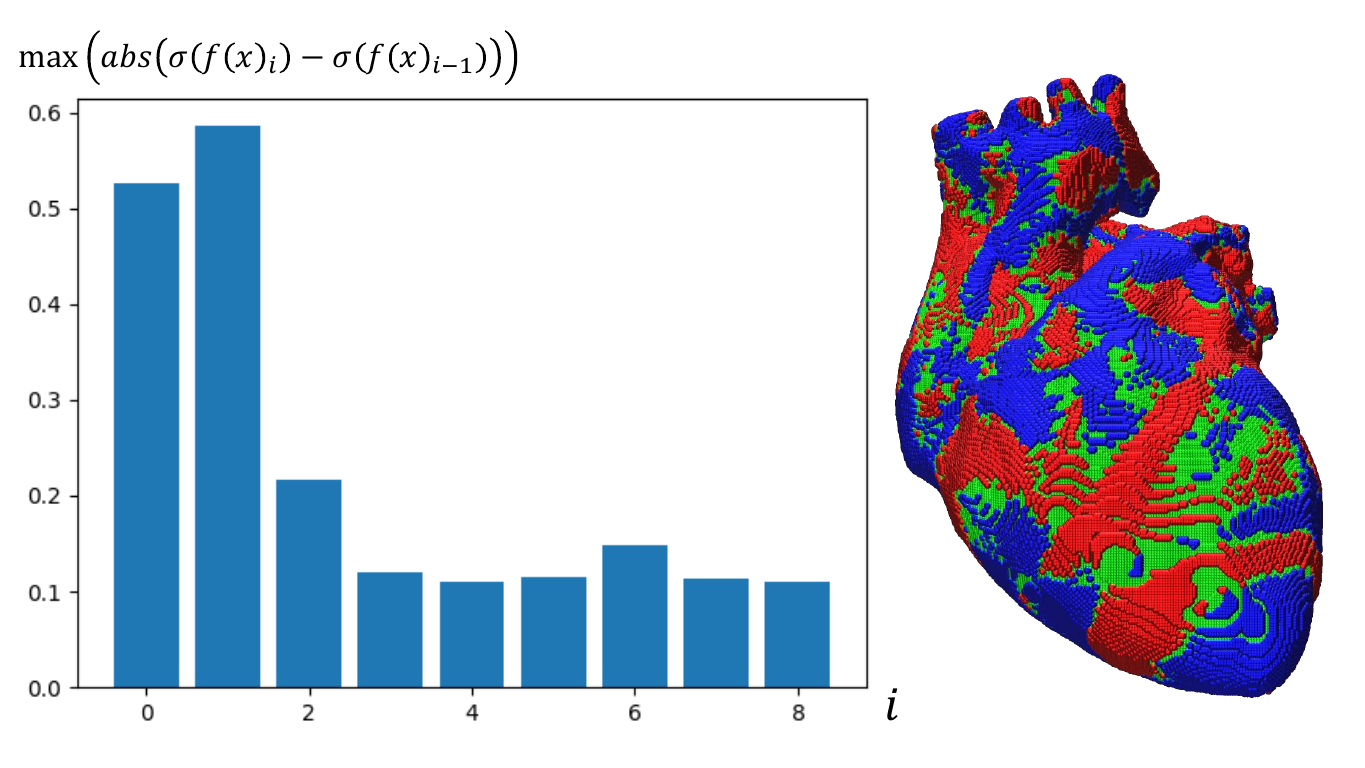}
    \caption{Refinement progression throughout the iteration process. The maximum change in the output value decreases along the iterations. Right: points that were added to the shape (Red) and that were removed from it (blue) in the last refinement iteration. These are concentrated on the level set, whereas most other points (green, covering all internal regions) have already converged.}
    \label{fig:restyle_ablation}
\end{figure}

OReXNet is simply an MLP that takes a 3D coordinate as input, represented using positional encoding, and produces the function $Y$  (Sec.~\ref{subsec:problem-setting}) at any query point. In order to encourage high-frequency details, we introduce an iterative refinement mechanism. Inspired by recent work  \cite{alaluf2021restyle}, we allow the network to refine its own results by running them through the network for a small number of iterations. This process was previously shown to produce a coarse-to-fine evolution in the realm of 2D images \cite{alaluf2021restyle}, and we argue it applies to our case as well. As demonstrated in Fig.~\ref{fig:pipeline}, ORexNet is hence a \textit{residual} MLP that is fed the result of the previous iteration, and a small hidden state code, along with the query point. These iterations ($10$ in our experiments) are performed both during training and inference; at inference time, only the last result is taken, and during training, the loss is applied to the results of all steps. The first iteration starts from a learned constant $C$. We show that this process indeed sharpens features and allows the incorporation of higher-frequency details in Fig.~\ref{fig:restyle_ablation}.

\subsection{Loss Function and Inference}
\label{subsec:loss-function-and-inference}

In order to present our loss function, we must first attend to an issue in the final stage of our pipeline. In this stage, we extract an explicit mesh for the boundary of the indicator function using Dual Contouring~\cite{ju2002dual}. This step uses a discretely sampled version of $Y(x)$ and $\nabla Y(x)$ 
on a regular 3D grid to extract the mesh.
As can be seen in Fig.~\ref{fig:hinge-loss} (and as witnessed other works \cite{erler2020points2surf}), this creates a ripples-like artifact, that correlates to the 3D grid resolution. This results from an aliasing of the sampling in these regions where the gradient magnitudes exhibit high variance. This effect could be mitigated via higher resolution grid sampling, which is expensive, or a post-process operation that may compromise the geometry. Instead, we propose revisiting the loss function of the training process, and incorporating regularization to reduce this sharp drop in value at the shape boundary. First, we omit the activation---in other words, we use the function $f(x)$ for contouring, while using $Y(x)$ only for training (Sec~\ref{subsec:problem-setting}). This de-radicalizes the function values, already offering a softer transition between inside and outside regions. In addition, we explicitly penalize strong transitions by using a \emph{hinge loss} (Eq.~\ref{eq:loss}). With this, we reduce the gradient magnitude variance considerably by limiting them close to the input shape boundary. Our loss function for a single point $x$ is then
\begin{equation}
\mathcal{L}(x, \theta) = {\sum_{i=0}^{N-1}{\text{BCE}(Y_i(x))}}+\lambda \max{(0,|| \nabla f_{N-1}(x) || - \alpha)}
\label{eq:loss}
\end{equation}
Where $\sigma$ is the sigmoid function, $\theta$ are the parameters of the network, $BCE()$ is the binary cross-entropy loss applied to all outputs of our iterative scheme, $max(0,x)$ is a hinge loss applied only to the last OReXnet iteration, $N$ the number of iterations, and $\alpha$ and $\lambda$ are hyperparameters. Fig.~\ref{fig:hinge-loss} demonstrates the correction effect of the hinge loss.

\section{Experiments}
\label{sec:results}

To evaluate our method, we developed a prototype, and ran it on a NVIDIA GeForce GTX 1080 Ti. See the Supplementary Material for exact timing and memory consumption statistics. Generally speaking, all training processes took less than four hours on the single GPU. 

\subsection{Results and Comparisons}

\begin{figure*}[htb]
    \centering
    \includegraphics[width=\textwidth]{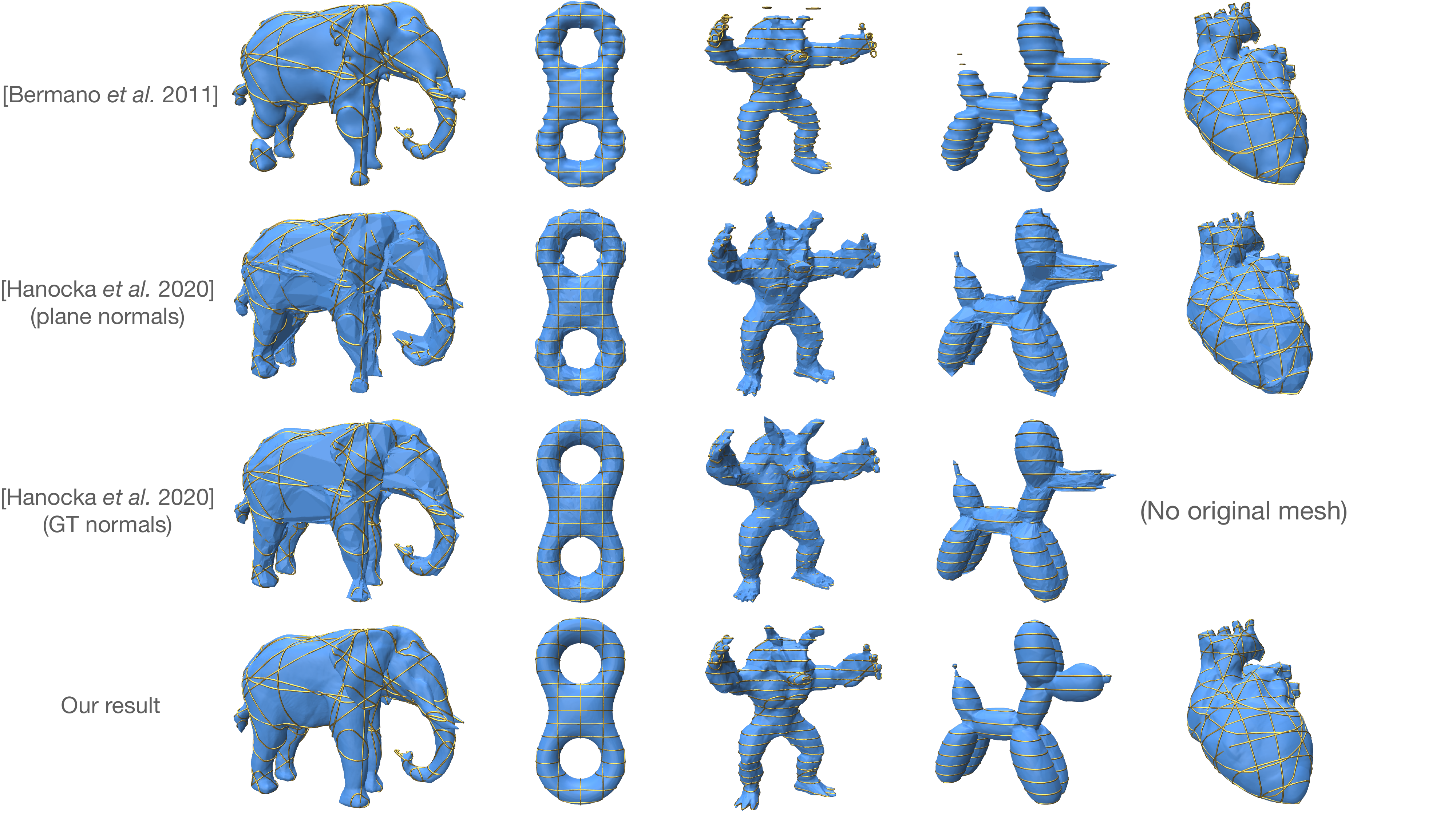}
    \caption{Qualitative comparisons to other approaches. Our result is smoother, more aligned to features, and with less artifacts. Note how the reconstructed shape silhouettes does not suffer imprints from the input planes. Check Table~\ref{tab:comparisons_table} for quantitative comparison.}
    \label{fig:comparisons_visual}
\end{figure*}

We first qualitatively demonstrate the result of our method on a variety of slice inputs, from both the medical and graphics worlds (Fig.~\ref{fig:gallery}). This demonstrates the versatility and generality of our method. We further show  our our algorithm correctly reproduces the internal cavities and details of reconstructed objects in Fig.~\ref{fig:xray}. Note how intricate details are learned along side a smooth interpolation between slices, leaving no slice transition artifacts on the resulting shape.

In terms of other methods, we qualitatively (Fig.~\ref{fig:comparisons_visual}) and quantitatively (Table~\ref{tab:comparisons_table}) compare our method to Bermano et al.~\cite{bermano2011online}, which is a reconstruction from cross-section method with the same input-output as ours. We further compare to general state-of-the-art reconstruction methods that target point clouds \cite{hanocka2020point2mesh}. Hanocka et al.~\cite{hanocka2020point2mesh} expect point normals as input as well. Hence we compare our result to the latter work with Ground Truth (GT) normals at input, and with ones coming from the input plane normals for a fairer comparison. We use the following metrics: 1) intersection-over-union volume of inside regions in reconstructed volumes and 2) GT planes, and 3) symmetric Hausdorff distance. These metrics demonstrate our superiority over the compared methods. See Supplementary Material for more comparisons.

\begin{figure}[ht]
    \centering
    \includegraphics[width=0.5\textwidth]{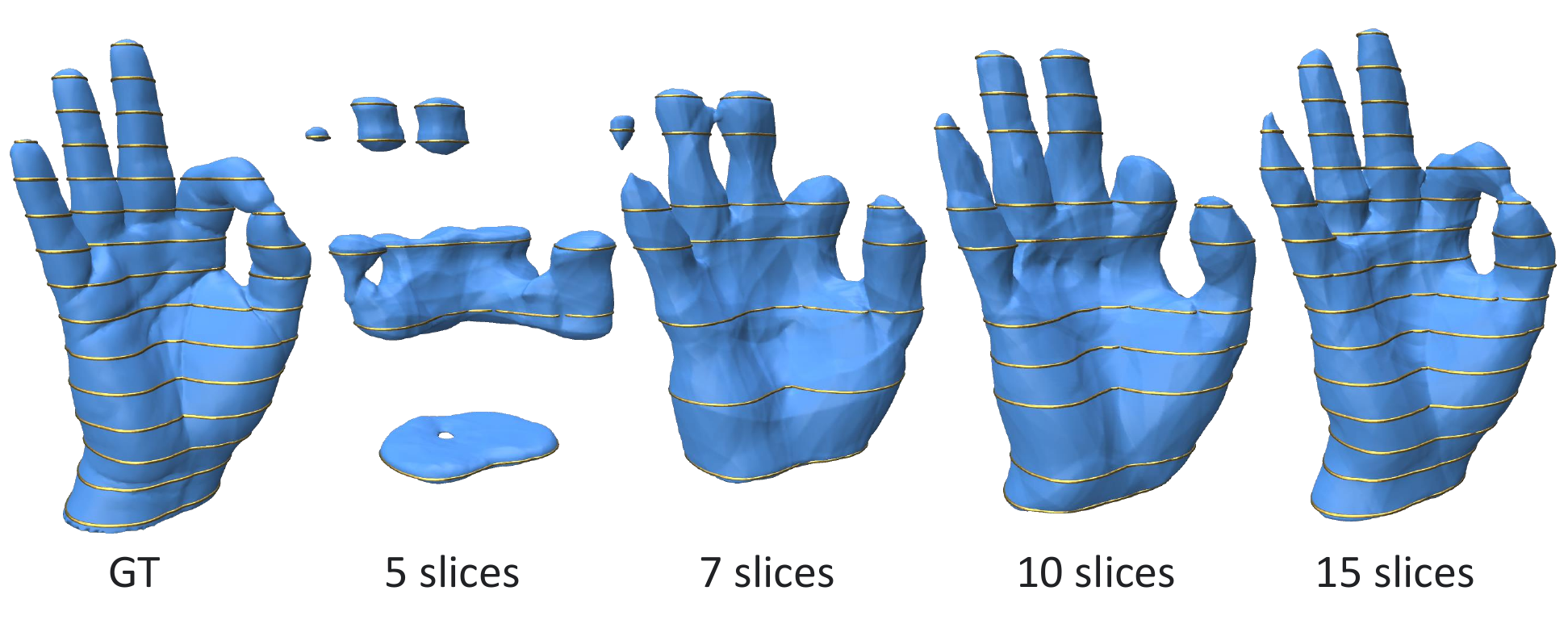}
    \caption{Increasing number of slices. As can be expected, introducing more slices converges consistently converges to the solution.}
    \label{fig:n-slices}
\end{figure}

\paragraph{Scalability} We show that our method does not suffer the computational cost that arrangement-based methods~\cite{bermano2011online,liu2008surface,huang2017topology} must bear, since we do not tessellate the space, and hence scale well with the input. In addition, we show (Fig.~\ref{fig:n-slices}) how our method converges with the addition of slices.

\subsection{Ablations}

We perform ablation testing to evaluate important aspects of our method and our design choices.

\begin{figure}[ht]
    \centering
    \includegraphics[width=0.47\textwidth]{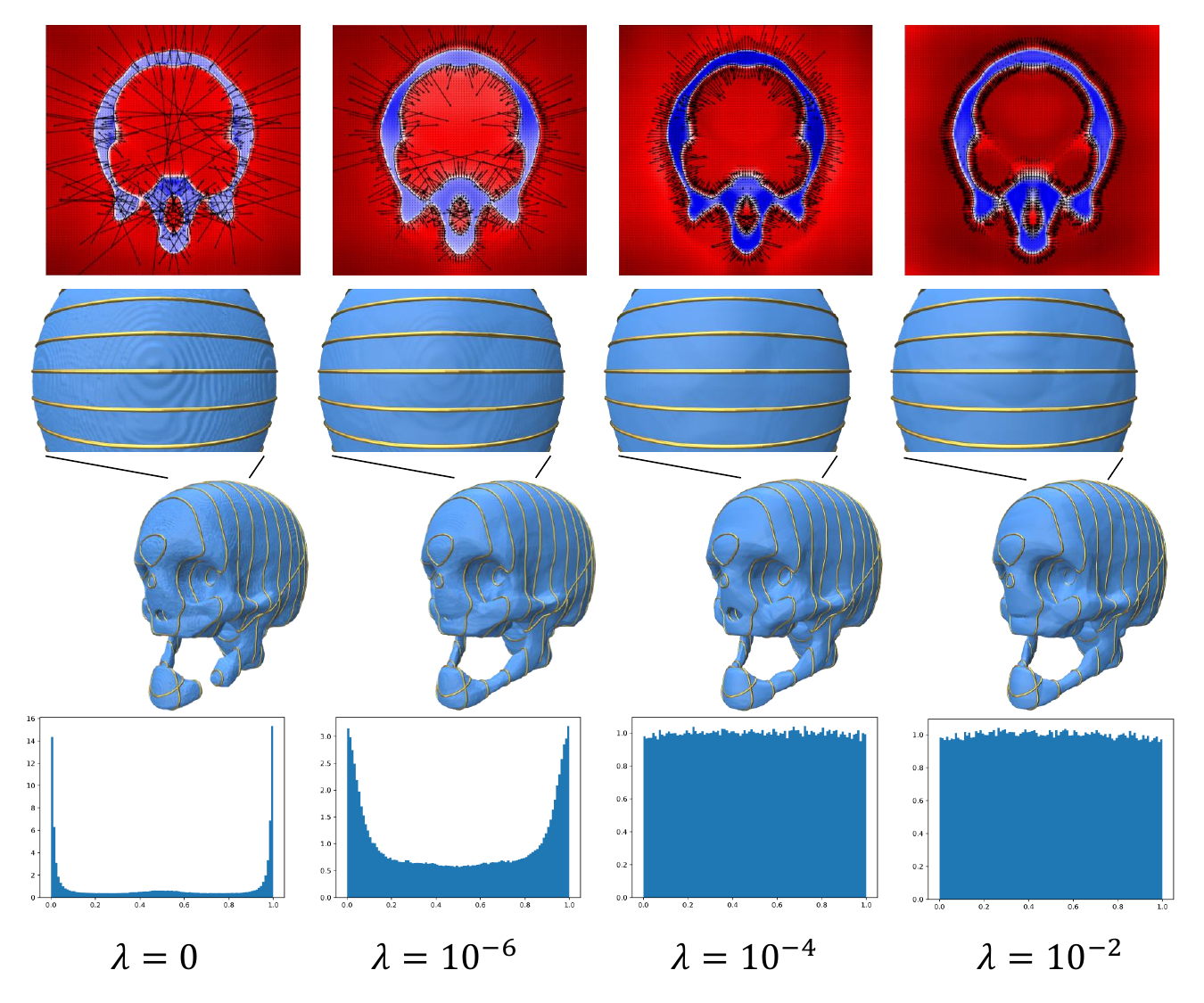}
    \caption{Left to right: our Hinge loss effect for mitigating ripple-like aliasing artifacts. Top: output values of $f(x)$ on a single slice (arrows show gradients). Middle: The reconstructed mesh with a closer view on the top of the skull. Bottom: distribution of vertices along grid edges during mesh extraction. It is evident the hinge loss effectively controls gradient magnitudes at the zero set, curbing the aliasing artifacts.  The hinge regularization further allows to place more vertices in the middle of the sampled grid cells, and not only on their edges.}
    \label{fig:hinge-loss}
\end{figure}

\paragraph{Gradient magnitude regularization (Fig.~\ref{fig:hinge-loss})} We show the effect of the hinge-loss regularizer on gradient magnitudes, with increasing $\lambda$ values. It is evident the hinge loss effectively filters out the high variance, and smooths the ripple artifacts. As can be seen (bottom row), without our hinge loss, geometry vertices tend to snap to discrete locations during the mesh extraction phase. This is because a large gradient induces a large bias in the magnitude of the indicator function, encouraging vertices to remain close to the grid points sampled during the extraction step. Regularizing the gradients around the boundry regions (top row) decreases the gap between the positive and negative sides of the function, allowing a more uniform, and hence continuous, distribution of the mesh vertices. With this bias removed, the resulting geometry is better sampled and hence the ripple artifact is mitigated (middle).

\begin{figure}[ht]
    \centering
    \includegraphics[width=0.47\textwidth]{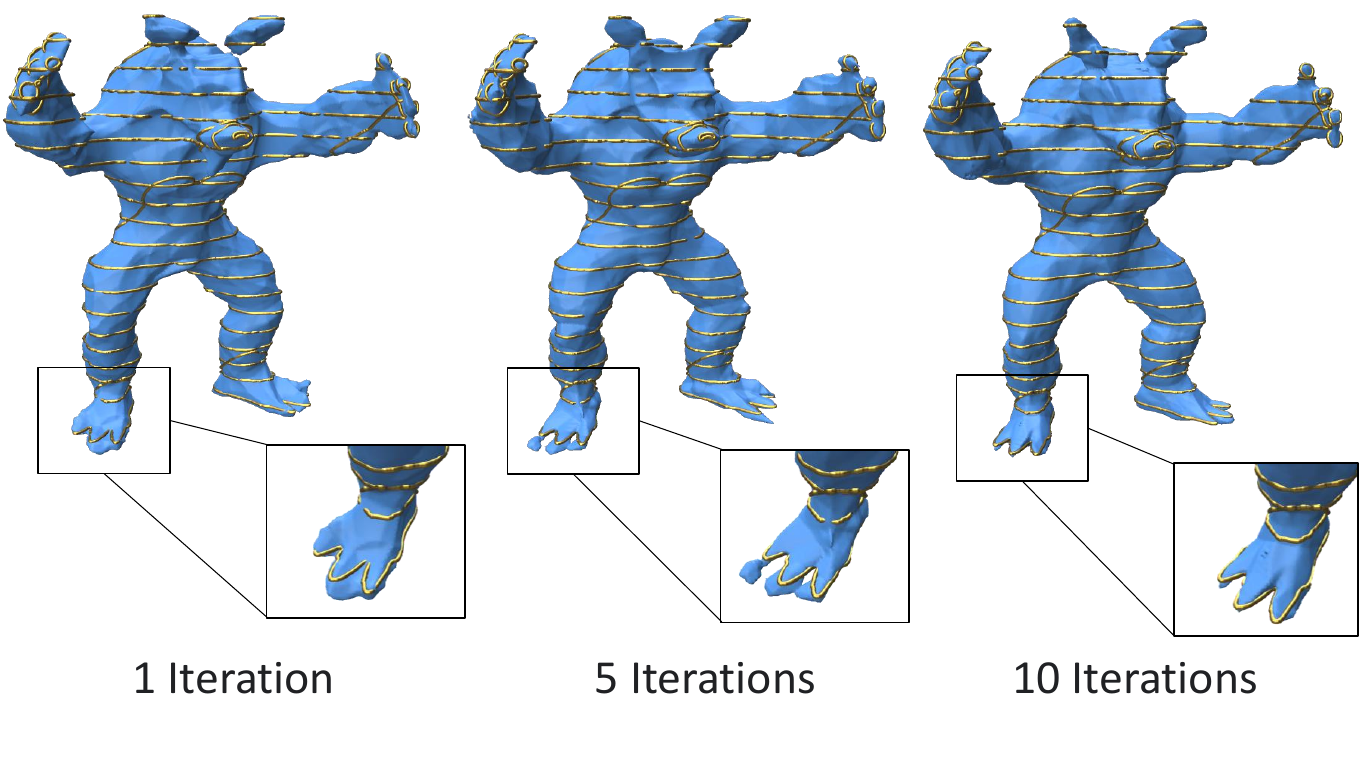}
    \caption{Iterative refinement experiment. Training and inferring with more OReXNet refinement iterations allows the network to perform smaller scale corrections and reduces spectral bias.}
    \label{fig:restyle-iterations}
\end{figure}

\paragraph{Iterative architecture (Fig.~\ref{fig:restyle-iterations})}
This ablations examines the effect of the number of refinement iterations on the accuracy of the mesh produced. As can be expected, the results show that as the number of iterations increases, high-frequency details are better represented.

\begin{figure}[ht]
    \centering
    \includegraphics[width=0.47\textwidth]{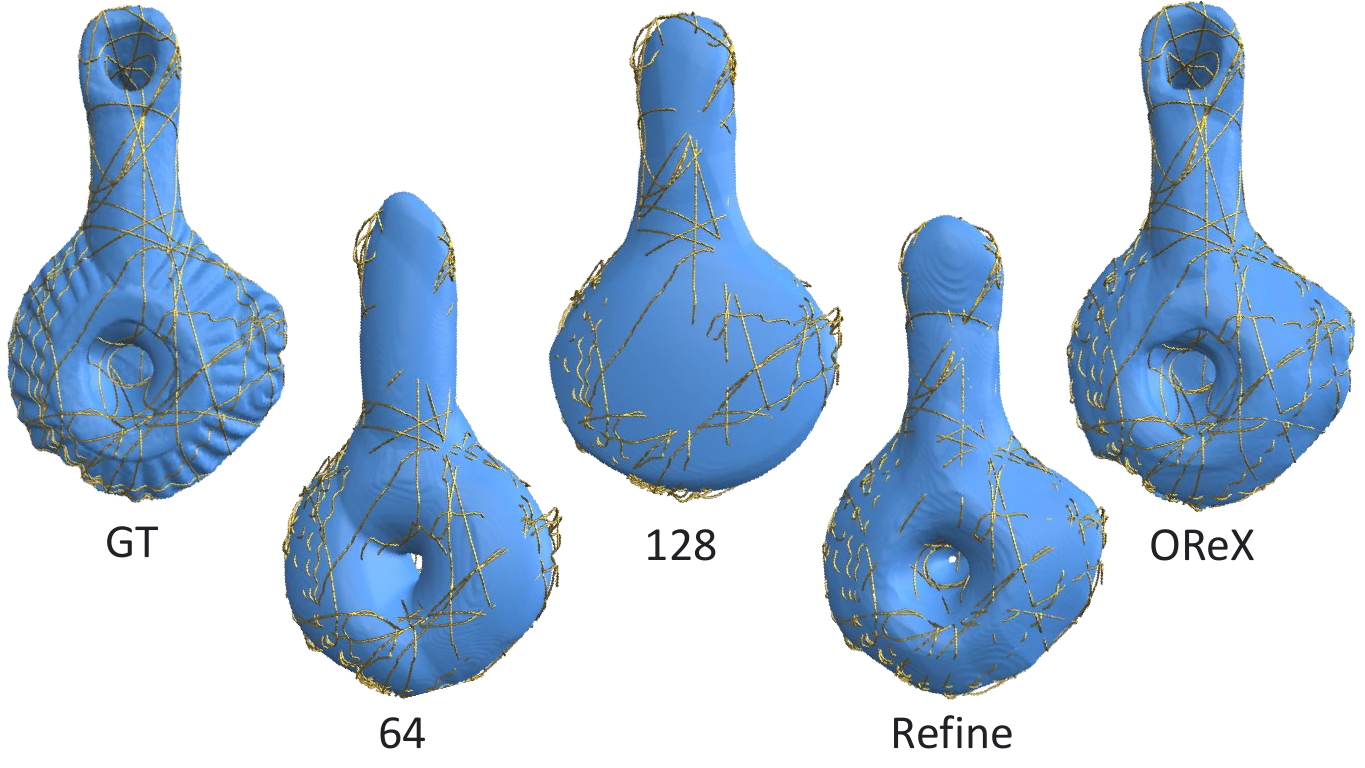}
    \caption{Sampling scheme Ablations. Left: ground truth. Middle: grid sampling with different resolutions, and a gradually increasing sampling resolution over the training process. Right: our method.}
    \label{fig:grid_sampling}
\end{figure}

\begin{figure}[ht]
    \centering
    \includegraphics[width=0.5\textwidth]{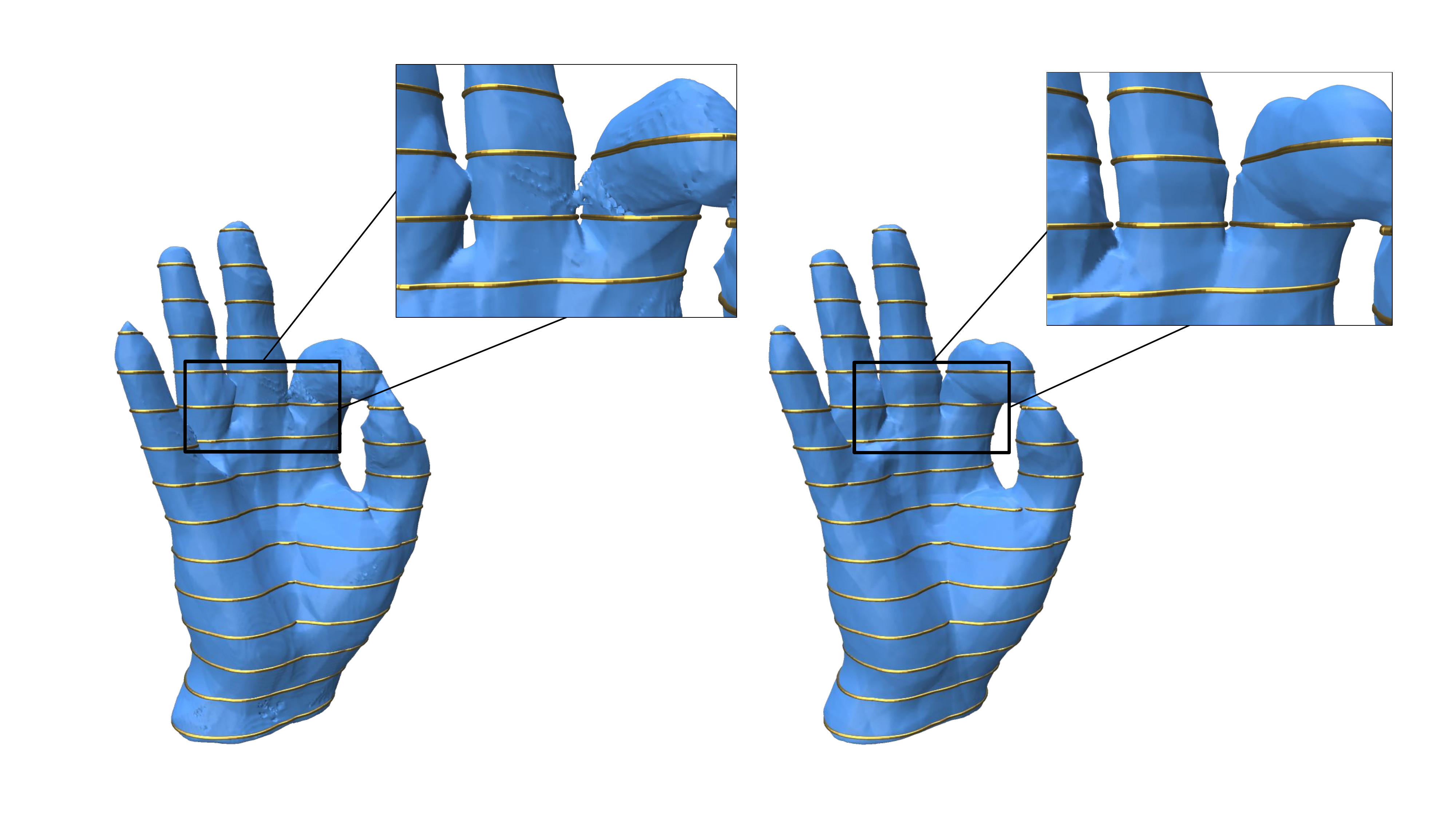}
    \caption{Left: non-hierarchical sampling, using points from all off-surface distances uniformly. Right: using our hierarchical sampling scheme.}
    \label{fig:no-refinment}
\end{figure}

\paragraph{Sampling method}
To evaluate our importance sampling scheme, we first replace it with regular grid sampling in two resolutions (Fig.~\ref{fig:grid_sampling}). As can be seen, using each of the resolutions alone yields overly smoothed results due to spectral bias. Training with a gradually increasing resolution allows better handling of high-frequency details. In addition, we have demonstrate the effects on quality when using our importance scheme in a non-hierarchical manner, i.e. using all points uniformly without consideration of distance from the boundary (Fig.~\ref{fig:no-refinment}). As can be seen, our scheme improves in detail quality upon both experiments,  demonstrating that our sampling scheme produces more accurate results with less spectral bias while using fewer samples.

\begin{figure}[ht]
    \centering
    \includegraphics[width=0.5\textwidth]{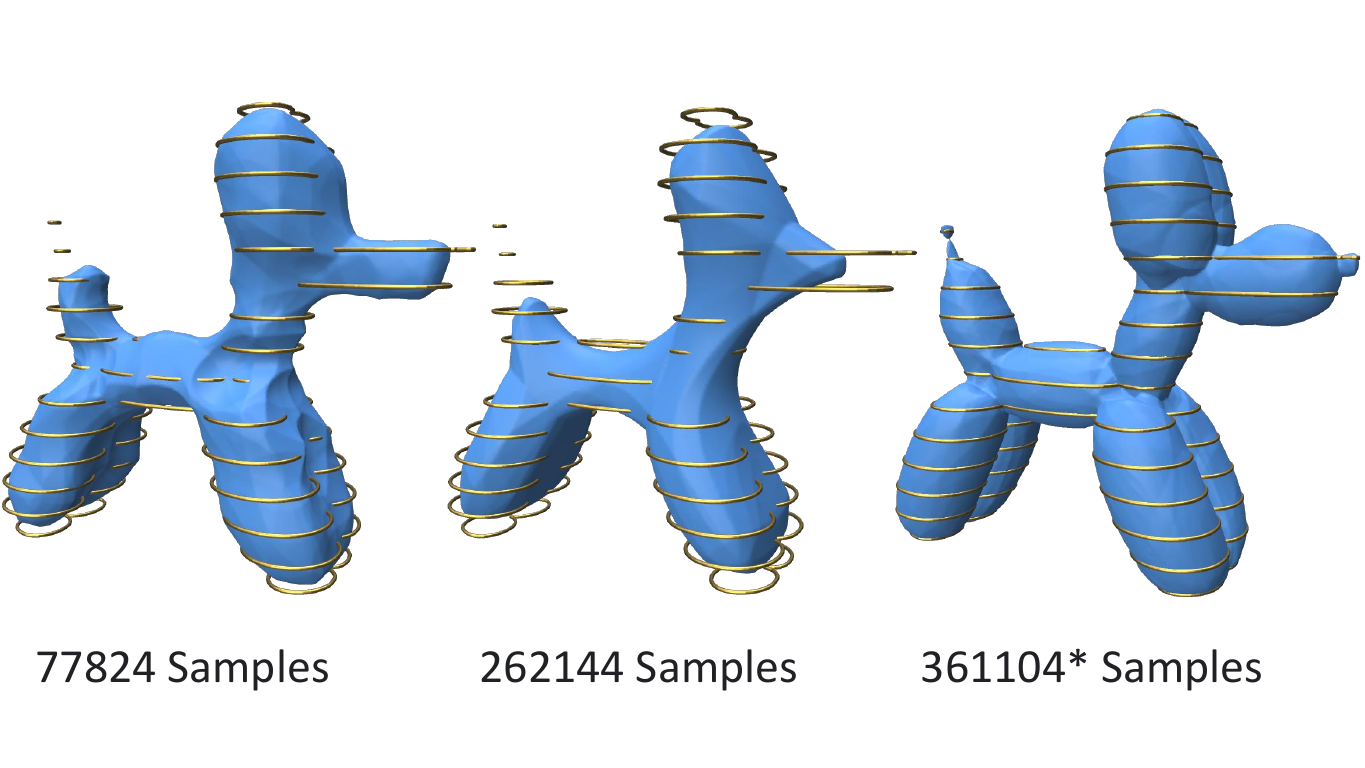}
    \caption{Left and middle: the baseline experiment with regular grid samples of 64 and 128 resolution over the slices. Right: our result for the same amount for samples (our variable number of samples is provided at its mean).}
    \label{fig:basic-nerf}
\end{figure}

\paragraph{Choice of architecture and sampling (Fig.~\ref{fig:basic-nerf})} We justify our design choices by training a baseline architecture with uniform grid sampling and no iteration refinement. It is evident such a model does not preserve details nor interpolate well, even for a relatively simple case.

\begin{table*}
  \centering
  \begin{adjustbox}{width=\textwidth}
  \begin{tabular}{|c || c c c c || c c c c || c c c c |}
    \toprule
    \multirow{3}{*}{Input} &
    \multicolumn{4}{c}{\underline{Hausdorff distance}} &
    \multicolumn{4}{c}{\underline{IoU in $3d$}} &
    \multicolumn{4}{c|}{\underline{IoU in $2d$}} \\
     & OReX & Bermano et. al & point2mesh  & point2mesh  & OReX & Bermano et. al & point2mesh  & point2mesh  & OReX & Bermano et. al  & point2mesh  & point2mesh  \\
     & &  & plane normals & GT normals & &  & plane normals & GT normals & &  & plane normals & GT normals\\
    \hline
    \midrule
    Eight (S) 
    & \textbf{0.018} & 0.065 & 0.219 & 0.046 
    & \textbf{0.984} & 0.865 & 0.842 & 0.961 
    & \textbf{0.988} & 0.984 & 0.795 & 0.980 \\
    
    Eight (M) 
    & \textbf{0.006} & 0.033 & 0.045 & 0.014 
    & \textbf{0.987} & 0.893 & 0.915 & 0.974 
    & \textbf{0.986} & 0.976 & 0.961 & 0.971 \\
    Elephant
    & \textbf{0.056} & 0.081 & 0.100 & 0.086
    & \textbf{0.966} & 0.908 & 0.885 & 0.935
    & \textbf{0.975} & 0.969 & 0.850 & 0.908 \\
    Balloon dog 
    & \textbf{0.049} & 0.194 & 0.078 & 0.086 
    & \textbf{0.957} & 0.868 & 0.897 & 0.928 
    & \textbf{0.988} & 0.977 & 0.926 & 0.956 \\
    Hand OK 
    & \textbf{0.063} & 0.177 & 0.195 & 0.135 
    & \textbf{0.955} & 0.860 & 0.921 & 0.931 
    & \textbf{0.987} & 0.968 & 0.908 & 0.882 \\
    Armadillo 
    & \textbf{0.050} & 0.121 & 0.057 & 0.059 
    & \textbf{0.939} & 0.891 & 0.891 & 0.921 
    & \textbf{0.964} & 0.776 & 0.850 & 0.868 \\
    \bottomrule
  \end{tabular}
  \end{adjustbox}
  \caption{Quantitative comparisons. We measure performance using Hausdorff distance, IoU of the inner volume compared to the GT shape, and IoU of the inner surface on the input cross-sections. We compare our result to a dedicated cross-section based reconstruction method \cite{bermano2011online}, and to two flavors of a pointcloud reconstruction method \cite{hanocka2020point2mesh}. See more comparisons in the Supplementary Material.}
  \label{tab:comparisons_table}
\end{table*}

\section{Conclusions}

We have presented OReX, a state-of-the-art method for the long-standing problem of shape reconstruction from planar cross-sectional indicator data. Free of datasets and training requirements, OReX is simple and intuitive to use. Our work balances the smoothness of a neural prior with high-frequency features. We show our approach successfully produces smooth interpolation between contours while respecting high-frequency features and repeating patterns. In addition, we believe some of the analysis and insights presented here can be applied to neural fields in similar tasks. 

The advantage of our method is also its disadvantage: using only binary data, much of the information is lost in the process. For example, for medical imaging an interesting future direction would be to directly use the raw grayscale density input of a planar probe such as that comes from an ultrasound.  Another interesting direction for research is extending this work for multi-labeled volumes (e.g., the reconstruction of several organs simultaneously from a scan), and using partial or noisy data. %

{\small
\bibliographystyle{ieee_fullname}
\bibliography{OReX_main/egbib}
}

\end{document}


\title{OReX: Object Reconstruction from Planar Cross-sections Using Neural Fields\newline
\textit{(Supplementary material)}}

\author{Haim Sawdayee\\
The Blavatnik School of Computer Science\\
Tel Aviv University\\
{\tt\small haimsawdayee@mail.tau.ac.il}
\and
Amir Vaxman\\
School of Informatics\\
The University of Edinburgh\\
{\tt\small avaxman@inf.ed.ac.uk}
\and
Amit H. Bermano\\
The Blavatnik School of Computer Science\\
Tel Aviv University\\
{\tt\small amberman@tauex.tau.ac.il}
}
\maketitle

We supplement and expand on details from our main document.

\section{Implementation details}

\subsection{Sampling and extraction}

We train our network for a total of 650 epochs. At the $0, 50, 100, 200, 300, 450$  epoch marks, we add new points to the dataset, where we always train on the last three such sets. The points are sampled from these distributions:
\begin{enumerate}
    \item We compute the convex hull of the object. We then scale it up by $5\%$ and sample $16384$ points on it uniformly by selecting faces with probability proportional to area and sampling a point inside it using triangle point picking method.
    \item We sample every plane uniformly with $2048$ points, within a bounding box that is aligned to the principal axes of the contours.
    \item For epoch $0$, we initially sample $n=2$ points uniformly on every edge, and off-surface points in distance $\epsilon=2^{-4}$. In subsequent epoch marks we add $n=2, 3, 3, 4, 5$ and $\epsilon=2^{-5}, 2^{-5}, 2^{-6}, 2^{-6}, 2^{-7}$.
\end{enumerate}

The mesh is extracted by dual-contouring on a grid sampled by inference of $f(x)$ (both values and gradients), where we use grid resolution $300^3$.

\subsection{Architecture}

Our base architecture is a $7$-hidden layer MLP of layer width $64$. Unlike a typical Neural Field, our MLP is residual, that is it computes an offset from the input, rather than the absolute values (as seen in Fig. 2 of the main paper). Input coordinates are modulated using $sin$ and $cos$ Positional Encoding, with $5$ learnable frequencies concatenated to the original input.

\textbf{Refinement iterations}: We run our prediction $10$ times through our network in a recurrent manner, both during training and test time. Except for the OReXNet output, the MLP also produces a small hidden code of size $32$ that is passed along the iterations to the network. The first iteration is fed with a learned constant $C$.

\subsection{Training hyperparameters}

For all experiments in the paper, we use the hinge loss with $\lambda=10^{-4}$ and set the hinge point to be at $\alpha=100$. We trained with initial learning rate $lr=10^{-2}$ and reduced it by a factor of $\gamma=0.9$ every $10$ epochs.

\subsection{Cross-section placement}

The placement and the density of the cross-section planes is a determining factor for resulting quality---in fact, this is the entire challenge, as regions that are not sampled can only be extrapolated, as in Fig~\ref{fig:missing-regions}.
To show the robustness of our method, our experiments show three patterns of selected cross-sections: randomly-oriented slices (Elephant, Oil lamp), regularly-spaced axis-aligned planes (Balloondog, Figure eight, Hand ok, Rose), and medically-generated slices (Skull, Hart, Vertebrate, Abdomen).

\section{Additional experiments}

\begin{figure}[h]
    \centering
    \includegraphics[width=0.5\textwidth]{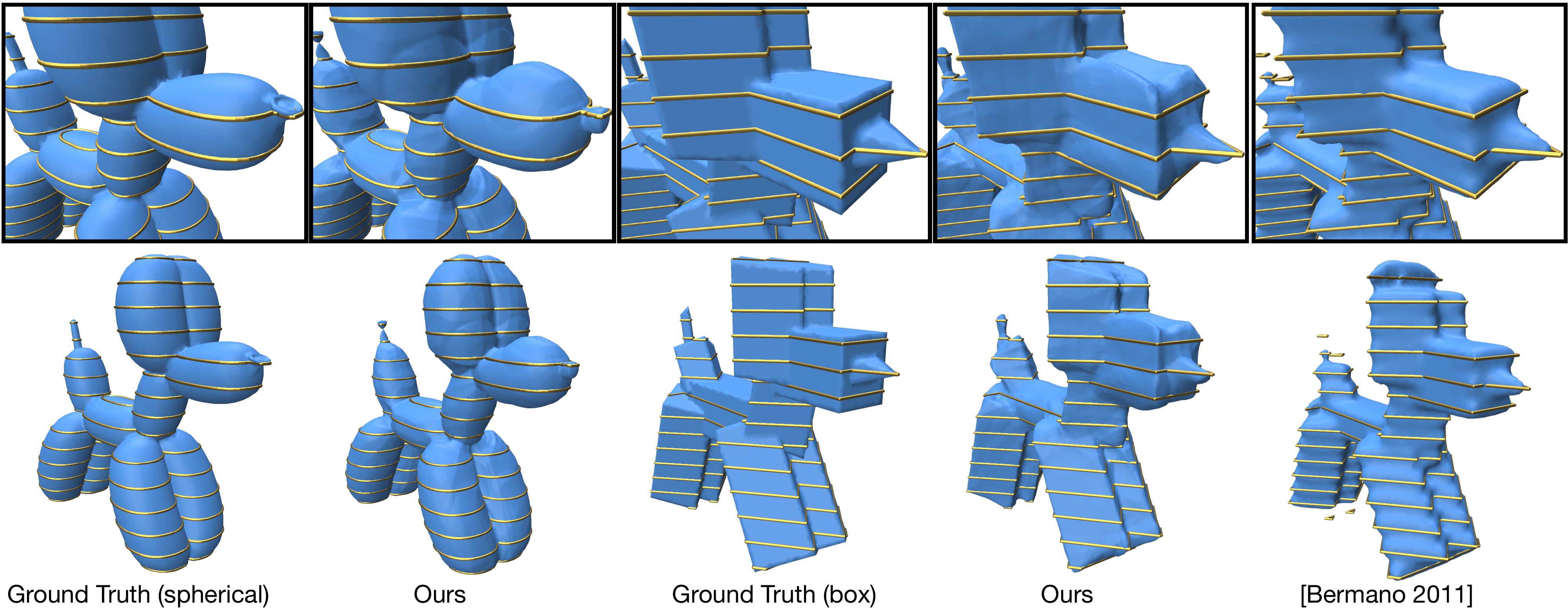}
    \caption{Self-similarities and how OReX leverages them. When the slices sample either a more round shape (left, 'round dog') or a cube-like sharp shape (middle 'box dog'), the interpolation of the nose is consistent with the rest of the style, thanks to the natural self-similarities of the neural field. Methods that use a smoothing prior (e.g.,~\cite{bermano2011online}, right) fail to capture the sharpness of the model.}
    \label{fig:balloondog-box}
\end{figure}

\begin{figure}[h]
    \centering
    \includegraphics[width=0.5\textwidth]{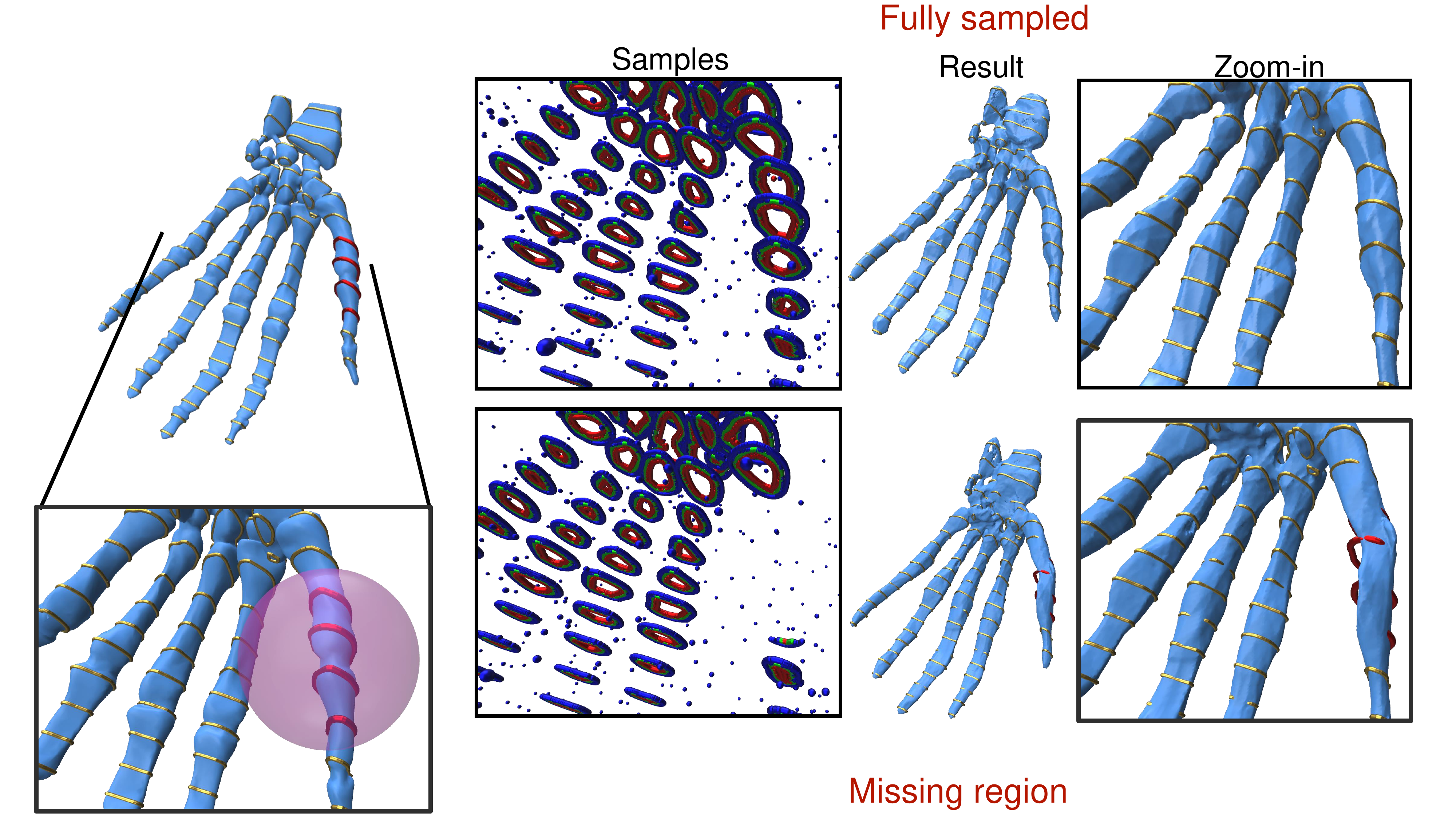}
    \caption{OReX naturally completes regions with missing samples, even as we omit them from the training process (red contours) Top row: the complete reconstruction. Bottom row: samples in the highlighted region are omitted. Everything else is trained as usual. Left to right: Grout Truth mesh, samples used in training, ORex reconstruction, with a zoomed-in view.}
    \label{fig:missing-regions}
\end{figure}

We demonstrate the advantages of using neural fields interpolators, rather than traditional smooth choices (like the mean-value interpolant by~\cite{bermano2011online}). In Fig~\ref{fig:balloondog-box} and Fig~\ref{fig:missing-regions} the neural field demonstrates self-similarity that correctly interpolates regions that are not sampled.

\section{Quantitative comparisons}

In Table~\ref{tab:comparisons_table_exp}, we expand our quantitative comparisonsquantitative comparisons (Table 1 in the main text). We compare our method to three others: \cite{bermano2011online} is a method developed specifically for cross-section reconstruction with similar settings. We attempted to run the code for newer methods \cite{zou2015topology,huang2017topology}, but could not produce results for or inputs. 

In addition, we compare to two point-cloud-based methods \cite{hanocka2020point2mesh, wang2021imls}, using our setting and with additional input information in the form of point normals. We measure performance using both global geometry measures (Hausdorf distance, Champfer Distance, and IoU of the reconstructed volumes), as well as metrics on the input cross-sections themselves (IoU in 2D). In all measures we demonstrate state-of-the-art results. In Fig ~\ref{fig:visual-dist} we visualized closest-point distances of our method.
        
\begin{table*}
  \centering
  \begin{adjustbox}{width=\textwidth}
  \begin{tabular}{|c || c c c c c || c c c c c || c c c c c|| c c c c c |}
    \toprule
    \multirow{3}{*}{Input} &
    \multicolumn{5}{c}{\underline{Hausdorff distance}} &
    \multicolumn{5}{c}{\underline{CD}} &
    \multicolumn{5}{c}{\underline{IoU in $3d$}} &
    \multicolumn{5}{c|}{\underline{IoU in $2d$}} \\
     &
     OReX & Bermano et. al & point2mesh  & point2mesh  & Neural-IMLS &
     OReX & Bermano et. al & point2mesh  & point2mesh  & Neural-IMLS &
     OReX & Bermano et. al & point2mesh  & point2mesh  & Neural-IMLS &
     OReX & Bermano et. al & point2mesh  & point2mesh  & Neural-IMLS \\
     & &  & plane normals & GT normals &
     & &  & plane normals & GT normals &
     & &  & plane normals & GT normals &
     & &  & plane normals & GT normals &\\
    \hline
    \midrule
    Eight 15 &
    \textbf{0.018} & 0.065 & 0.219 & 0.046 & 0.237 & 
    \textbf{0.002} & 0.020 & 0.029 & 0.005 & 0.035 & 
    \textbf{0.984} & 0.865 & 0.842 & 0.961 & 0.680 & 
    \textbf{0.988} & 0.984 & 0.795 & 0.980 & ** \\
    Eight 20 &
    \textbf{0.006} & 0.033 & 0.045 & 0.014 & 0.052 & 
    \textbf{0.002} & 0.016 & 0.013 & 0.004 & 0.004 & 
    \textbf{0.987} & 0.893 & 0.915 & 0.974 & 0.971 & 
    \textbf{0.986} & 0.976 & 0.961 & 0.971 & 0.975 \\
    Elephant & 
    \textbf{0.056} & 0.081 & 0.100 & 0.086 & 0.312 & 
    \textbf{0.006} & 0.015 & 0.018 & 0.010 & 0.031 & 
    \textbf{0.966} & 0.908 & 0.885 & 0.935 & 0.803 & 
    \textbf{0.975} & 0.969 & 0.850 & 0.908 & ** \\
    Balloon dog &
    \textbf{0.049} & 0.194 & 0.078 & 0.086 & 0.264 & 
    \textbf{0.006} & 0.021 & 0.015 & 0.010 & 0.044 & 
    \textbf{0.957} & 0.868 & 0.897 & 0.928 & 0.659 & 
    \textbf{0.988} & 0.977 & 0.926 & 0.956 & ** \\
    hand ok &
    \textbf{0.063} & 0.177 & 0.195 & 0.135 & 0.207 & 
    \textbf{0.008} & 0.024 & 0.015 & 0.013 & 0.040 & 
    \textbf{0.955} & 0.860 & 0.921 & 0.931 & 0.707 & 
    \textbf{0.987} & 0.968 & 0.908 & 0.882 & 0.765 \\
    Armadillo &
    \textbf{0.050} & 0.121 & 0.057 & 0.059 & 0.827 & 
    \textbf{0.009} & 0.017 & 0.016 & 0.011 & 0.212 & 
    \textbf{0.939} & 0.891 & 0.891 & 0.921 & * & 
    \textbf{0.964} & 0.776 & 0.850 & 0.868 & * \\
  
    \bottomrule
  \end{tabular}
  \end{adjustbox}
  \caption{Quantitative comparisons. We measure performance using the global metrics Hausdorff distance, Champfer Distance, and Intersection over Union (IoU) of the inner volume compared to the GT shape. In addition, we measure fidelity performance by reporting IoU of the ``inside'' regions on the input cross-sections. We compare our result to a dedicated cross-section-based reconstruction method \cite{bermano2011online}, to two flavors of a point-cloud reconstruction method \cite{hanocka2020point2mesh}, and another recent point-cloud based method \cite{wang2021imls}. }
  \label{tab:comparisons_table_exp}
\end{table*}

\section{Qualitative comparisons}
\begin{figure}[h]
    \centering  
    \includegraphics[width=0.5\textwidth]{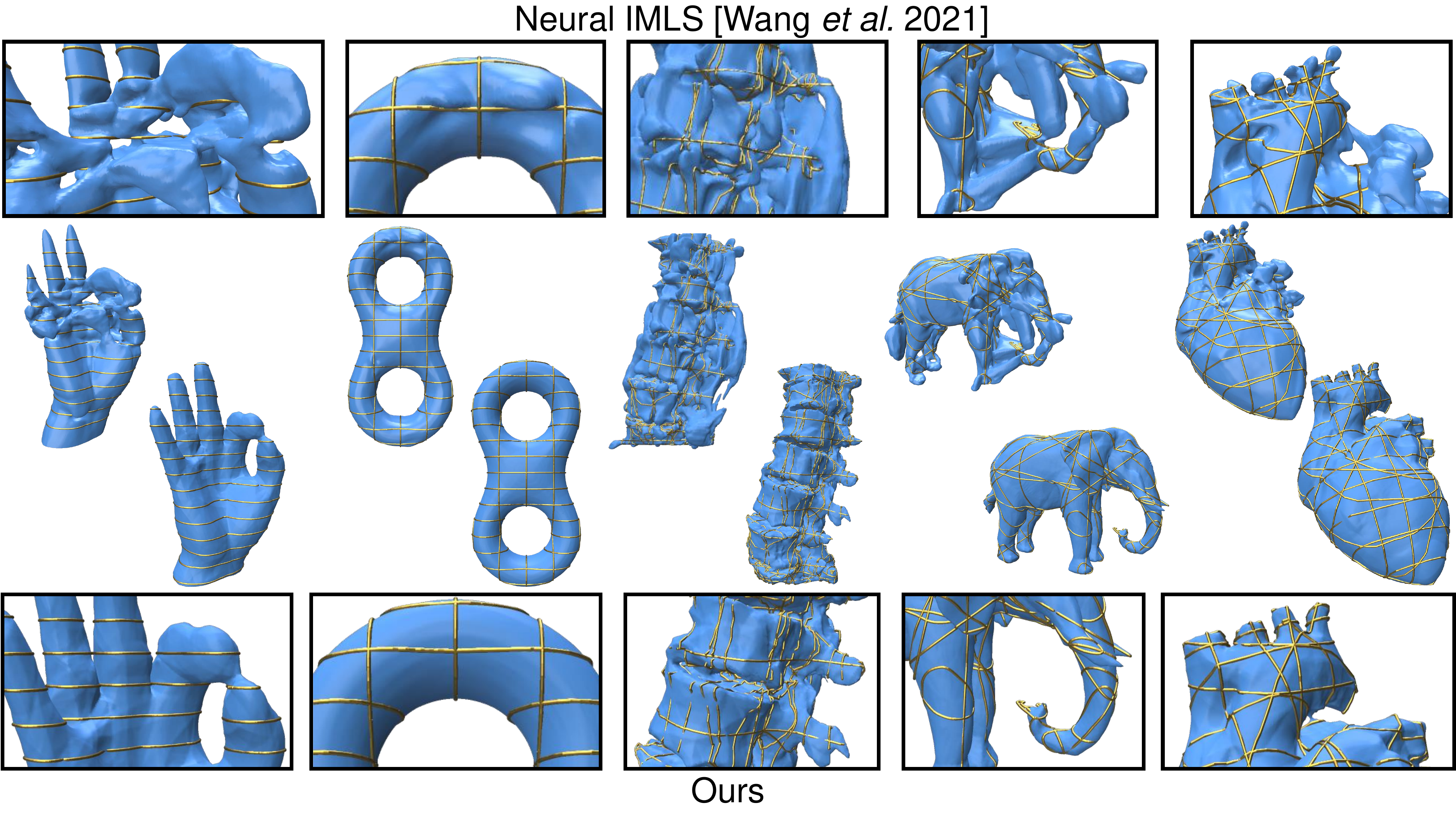}
    \caption{Comparison of OReX (bottom) to a recent point-cloud-based method \cite{DBLP:journals/corr/abs-2109-04398} (top).  Point-cloud methods typically expect a dense input covering the whole volume more or less uniformly; hence artifacts appear when feeding cross-sectional input.}
    \label{fig:IMLS}
\end{figure}

\begin{figure}[h]
    \centering
    \includegraphics[width=0.5\textwidth]{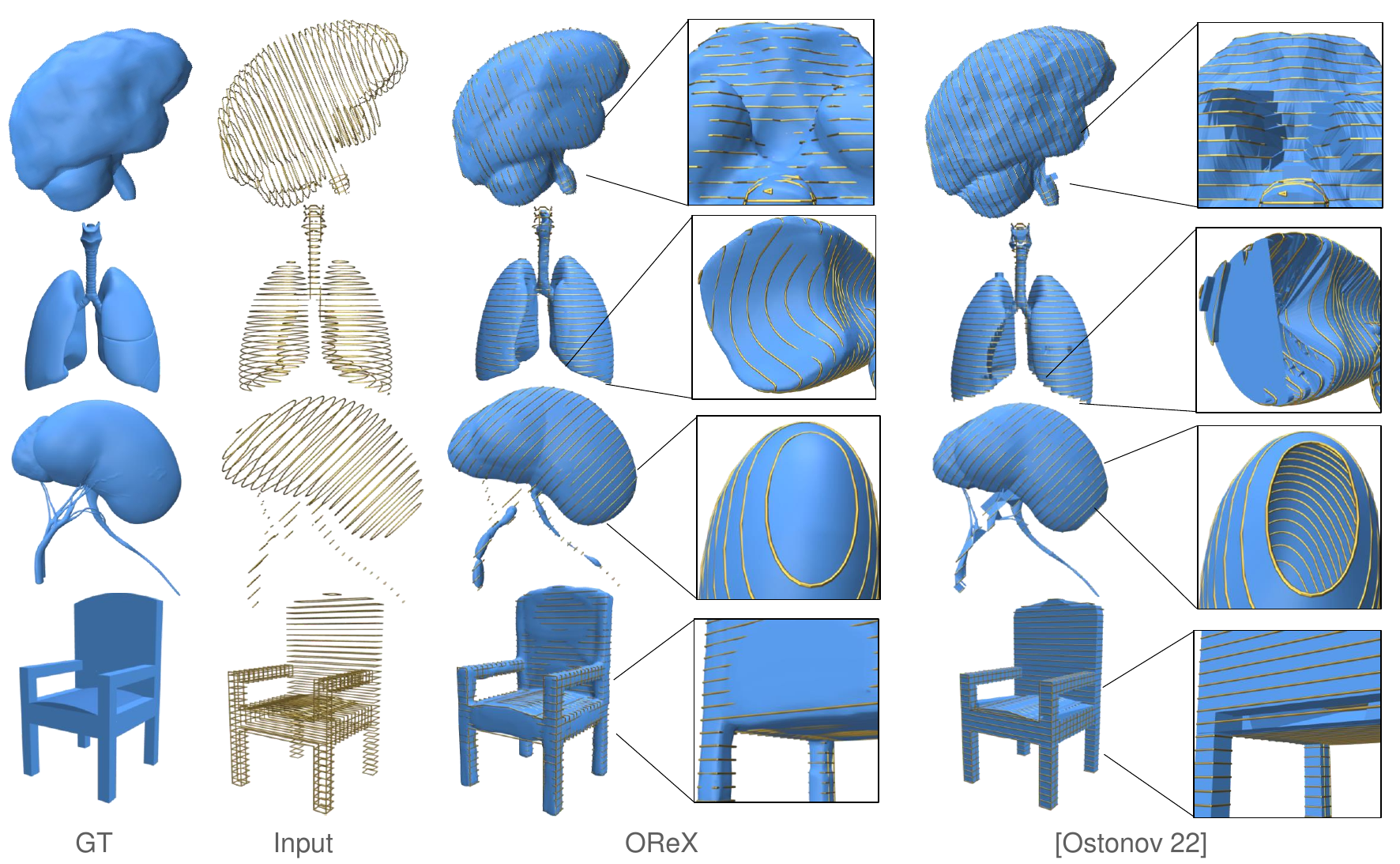}
    \caption{Comparison of OReX to a concurrent work by Ostonov ~\cite{ostonov2022cut}. Left to right: ground truth, input slices, our method, and Ostonov 22.}
    \label{fig:Ostonov}
\end{figure}

\begin{figure}[ht]
    \centering
    \includegraphics[width=0.5\textwidth]{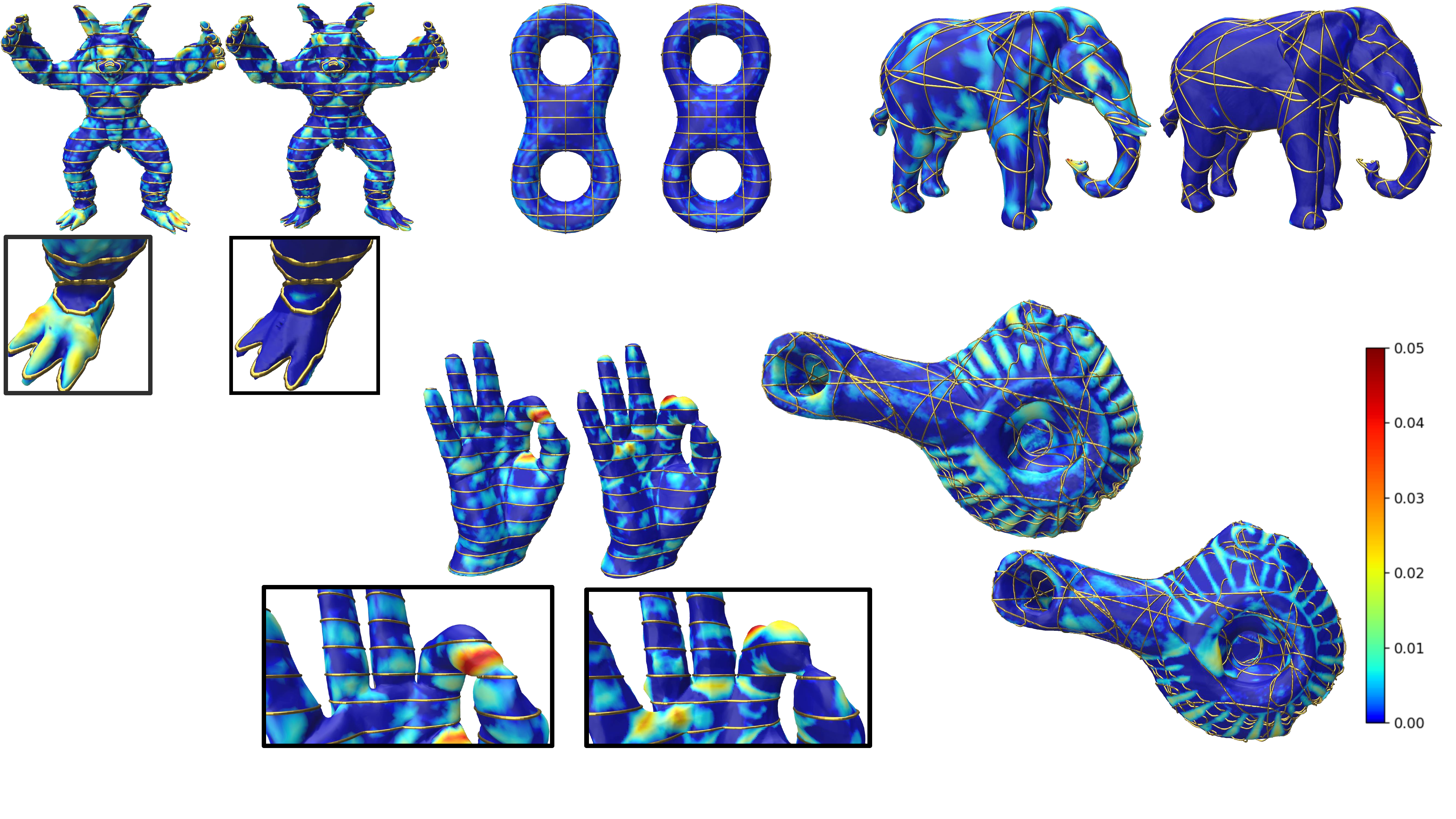}
    \caption{Closest-point distances from the ground truth (left
    meshes) to our reconstruction, and vice versa (right meshes). Scale
    is relative to the bounding-box diagonal.}
    \label{fig:visual-dist}
\end{figure}

Below are further visual comparisons with a recent point cloud method (fig.~\ref{fig:IMLS}) and a concurrent reconstruction method based on slices (fig.~\ref{fig:Ostonov}).
Our method produces watertight meshes that smoothly interpolate the input.

\section{Stats}

In Table~\ref{tab:stats}, we report statistics on the inputs we used. Note the complexity of the shape as well as runtimes.  

\begin{table*}[ht]
  \centering
  \begin{adjustbox}{width=\textwidth}
  \begin{tabular}{|c || c c c || c c |}
    \hline
    Input & \#slices & \#edges & \#samples (last set) & Training time (h) & Meshing time (h) \\
    \hline
    Armadillo & 26 & 11039 & 290412 & 2.90 & 0.22 \\
    Balloon dog & 15 & 4579 & 138684 & 1.51 & 0.22 \\
    Eight (S) & 15 & 2028 & 87664 & 1.06 & 0.22 \\
    Eight (M) & 20 & 3178 & 120904 & 1.40 & 0.21 \\
    Elephant & 24 & 10429 & 274116 & 2.75 & 0.23 \\
    Hand OK & 15 & 4082 & 128744 & 1.42 & 0.22 \\
    Oil lamp & 34 & 17387 & 433756 & 4.13 & 0.22 \\
    Abdomen & 42 & 53448 & 1067530 & 10.62 & 0.23 \\
    Heart & 25 & 4534 & 158264 & 1.77 & 0.23 \\
    Horse & 29 & 4101 & 157796 & 1.83 & 0.23 \\
    Twisted rose & 15 & 9000 & 227104 & 2.26 & 0.27 \\
    Skull & 16 & 8821 & 225572 & 2.31 & 0.33 \\
    Vertebrae & 36 & 14844 & 386847 & 4.22 & 0.24 \\  
    \hline
  \end{tabular}
  \end{adjustbox}
  \caption{Summary of our model zoo statistics. We report the number of planar cross-sections (\#slices), contour complexity (i.e., the total number of edges of the slices) (\#edges), the number of points sampled on the last set, closest to the contour (\#samples), training time (in hours), and time to extract the mesh after training (meshing time, in hours)}
  \label{tab:stats}
\end{table*}

{\small
\bibliographystyle{ieee_fullname}
\bibliography{OReX_sup/egbib}
}